\documentclass[11pt]{article}
\usepackage{arxiv}
\usepackage{times}
\usepackage{url}
\usepackage{latexsym}

\title{hinglishNorm - A Corpus of Hindi-English Code Mixed Sentences for Text Normalization}

\author{
 Piyush Makhija \\
   \And
  Ankit Kumar\\
  \AND
  Anuj Gupta
 \AND
 Vahan Inc\\
 firstname@vahan.co\\
}
\date{}

\begin{document}
\maketitle
\begin{abstract}
We present \emph{hinglishNorm} - a human annotated corpus of Hindi-English code-mixed sentences for text normalization task. Each sentence in the corpus is aligned to its corresponding human annotated normalized form. To the best of our knowledge, there is no corpus of Hindi-English code-mixed sentences for text normalization task that is publicly available. Our work is the first attempt in this direction. The corpus contains 13494 parallel segments. Further, we present baseline normalization results on this corpus. We obtain a Word Error Rate (WER) of 15.55, BiLingual Evaluation Understudy (BLEU) score of 71.2, and Metric for Evaluation of Translation with Explicit ORdering (METEOR) score of 0.50.

\end{abstract}

\section{Introduction}
Hindi is the fourth most-spoken first language in the world\footnote{https://en.wikipedia.org/wiki/List\_of\_languages\newline\_by\_number\_of\_native\_speakers}. According to one estimate, nearly 0.615 billion people speak Hindi as their first language\footnote{https://blog.busuu.com/most-spoken-languages\-in\-the\-world/}. Of these, most of the speakers are in India. The second most spoken language in India is English\footnote{https://en.wikipedia.org/wiki/List\_of\_languages\newline\_by\_number\_of\_native\_speakers\_in\_India}. Hindi and English are the official languages of the Indian Commonwealth\footnote{https://en.wikipedia.org/wiki/Hindi}. A large number of these people have joined the Internet recently.  As a matter of fact, Next Billion Users (NBU) is a term commonly used in tech and business circles to refer to the large number of people from India, Brazil, China and South-East Asia who joined the Internet in the last decade\footnote{https://www.blog.google/technology/next\-billion\-users/next\-billion\-users\-are\-future\-internet/}. This phenomena is primarily attributed to ubiquitous highly affordable phone and internet plans\footnote{https://www.hup.harvard.edu/catalog.php?isbn=9780674983786}. A large fraction of NBU users come from India and speak Hindi as either their first or second language. A large number of these people use a blend of Hindi and English in their daily informal communication. This hybrid language is also known as Hinglish\footnote{https://en.wikipedia.org/wiki/Hinglish}.

These users extensively use Internet platforms for User Generated Contents (UGC) - social media platforms such as Facebook or Twitter; messaging platforms such as WhatsApp or Facebook messenger; user reviews aggregators such as Google play store or Amazon. A key characteristic of their behaviour on such platforms is their use of Hinglish. Thus, building any Natural Language Processing (NLP) based Internet applications for these users necessitates the ability to process this `new' language. Further, these UGC platforms are notoriously noisy. This means there is an additional challenge of non-canonical text. Therefore, a key step in building applications for such text data is \textit{text normalization}. Intuitively, it is transforming text to a form where written text aligned to its normalized spoken form\cite{sproat2016rnn}. More formally, \textit{it is the task of mapping non-canonical language, typical of speech transcription and computer-mediated communication, to standardized writing}\cite{lusetti2018encoder}. 

Separately, there has been a lot of work in the two areas of normalization and building corpora of Hindi-English code mix text data, not much has been done at the intersection of the two(refer to section 2). To the best of our knowledge, there does not exist a corpus of Hindi-English Code Mixed sentences for normalization where the normalizations are human annotated. This work is an effort to release such a corpus

This work is motivated from our business use case where we are building a conversational system over WhatsApp to screen candidates for blue-collar jobs. Our candidate user base often comes from tier-2 and tier-3 cities of India. Their responses to our conversational bot are mostly a code mix of Hindi and English coupled with non-canonical text (ex: typos, non-standard syntactic constructions, spelling variations, phonetic substitutions, foreign language words in non-native script, grammatically incorrect text, colloquialisms, abbreviations, etc). The raw text our system gets is far from clean well formatted text and text normalization becomes a necessity to process it any further. 

The main contributions of this work are two-fold, viz. (i) creating a human annotated corpus for text normalization of Hindi-English code mix sentences; and (ii) reporting baseline metrics on the corpus. Further, we release the corpus and annotations under a Creative Commons Attribution-NonCommercial-ShareAlike License\footnote{http://creativecommons.org/licenses/by-nc-sa/4.0/}.

\section{Related Work}

In this section, we present relevant work in the following areas viz.(1) Text Normalization (2) Normalization and UGC Datasets (3) Code-mixed Datasets, (4) Hindi-English Datasets.

\noindent \textbf{Text Normalization}: Text normalization, sometimes also called lexical normalization, is the task of translating/transforming a non-standard text to a standard format. Using text normalization on noisy data, one can provide cleaner text data to downstream NLP tasks and improve the overall system performance\cite{liu2012broad}\cite{satapathy2017phonetic}. Some of the early work used a rule-based spell-checker approach to generate a list of corrections for any misspelled word, ranked by corresponding posterior probabilities\cite{church1991probability}\cite{mays1991context}\cite{brill2000improved}. However, this approach did not factor in any context while normalizing words. \cite{choudhury2007investigation} used a Hidden Markov Model (HMM), where they modeled each standard English word as a HMM and calculated  the  probability  of  observing  the  noisy token. “Moses”, a well known Statistical Machine Translation (SMT) tool, provided significant improvements in comparison to previous solutions\cite{koehn2007moses}. \cite{aw2006phrase} adapted a phrase-based Machine Translation (MT) model for  normalizing  SMS  and  achieved  significant gain in  performance. In the past few years, Neural network based approaches for text normalization have become increasingly popular and have shown competitive performance in shared tasks\cite{chrupala2014normalizing}\cite{min2015ncsu_sas_wookhee}. \cite{lusetti2018encoder}, \cite{liu2012broad} and \cite{satapathy2017phonetic} provide excellent literature covering the landscape on this topic.

\noindent \textbf{Normalization and UGC Datasets}: \cite{han2011lexical} introduced a text normalization approach for twitter data using a variety of supervised \& unsupervised learning techniques. This study resulted in `lexNorm'\footnote{http://people.eng.unimelb.edu.au/tbaldwin/etc/lexnorm\_v1.2.tgz}, an open-source dataset containing 549 tweets. \cite{baldwin2015shared} subsequently released lexNorm15\footnote{https://github.com/noisy-text/noisy-text.github.io/blob/master/2015/files/lexnorm2015.tgz}. This new dataset contained 2950/1967 annotated tweets in train/test sets. \cite{michel2018mtnt} created the MTNT dataset\footnote{https://www.cs.cmu.edu/~pmichel1/mtnt/} containing translations of Reddit comments from the English language to French/ Japanese and  vice versa, containing 7k$\sim$37K data points per language pair. This dataset contains user-generated text with different kinds of noise, e.g., typos, grammatical errors, emojis, spoken languages, etc. for two language pairs. \cite{van2017monoise} introduced ‘MoNoise’, a general purpose model for normalizing UGC text data. This model utilizes Aspell spell checker, an n-gram based language model and word embeddings trained on a few million tweets. It gave significant improvement in State-Of-The-Art (SOTA) normalization performance on the lexNorm15 dataset.  \cite{muller2019enhancing} focused on enhancing BERT model on UGC by applying lexical normalization. 

\noindent \textbf{Code-Mixed Datasets}: Since the launch of EMNLP  Shared  Tasks  of  Language  identification  in  Code-Switched Data\footnote{http://emnlp2014.org/workshops/CodeSwitch/call.html}, there has been an increased focus on analyzing the nature of code-mixed data, language identification approaches and how to carry out NLP tasks like POS tagging and Text normalization on such text data. For the first shared task, code-switched data was collected for language pairs such as Spanish-English (ES-EN),  Mandarin-English  (MAN-EN), Nepali-English(NEP-EN) and Modern Standard Arabic - Dialectal Arabic(MSA-DA)\cite{solorio2014overview}. Subsequently more language pairs were added with primary focus on language identification task\cite{molina2019overview}. \cite{aguilar2019named} introduced Named Entity Recognition on Code-Switched Data. \cite{mandal2018preparing} introduced Bengali-English code-mixed corpus for sentiment analysis. More recently, normalization of code mixed data has been receiving a lot of attention. \cite{barik2019normalization} worked on normalizing Indonesian-English code-mixed noisy social media data. Further, they released 825 annotated tweets from this corpus\footnote{https://github.com/seelenbrecher/code-mixed-normalization/tree/master/data }. \cite{phadte2017towards} focused on normalization of Konkani-English code-mixed text data from social media. 

\noindent \textbf{Hindi-English Datasets}: \cite{vyas2014pos} was one of the earliest work to focus on creating a Hindi-English code-mixed corpus from social media content for POS tagging. The same year \cite{bali2014borrowing} analyzed Facebook English-Hindi posts to show a significant amount of code-mixing. \cite{bhat2018universal} worked with similar English-Hindi code-mixed tweets in roman script for dependency parsing. \cite{patra2018sentiment} worked on sentiment analysis of code mixed Hindi-English \& Bengali-English language pairs. \cite{singh2018automatic} focused on normalization of code-mixed text using pipeline processing to improve the performance on POS Tagging task. indicnlp\_catalog\footnote{https://github.com/anoopkunchukuttan/indic\_nlp\_library} is a effort to consolidate resources on Indian languages. Table~\ref{table:hiEndatasets} presents the most relevant Hindi-English datasets from this effort. 

\begin{table}
\centering
\begin{tabular}{p{0.55\columnwidth}|p{0.2\columnwidth}|p{0.2\columnwidth}}
\hline
\textbf{}  \textbf{Dataset} & \textbf{Task} & \textbf{Size} \\
\hline 
IITB English-Hindi Parallel Corpus \cite{anoop2018iit} & Machine Translation
 & Train - 1,561,840 \newline Dev - 520 \newline Test - 2,507  \\
 & & \\
HindiEnCorp 0.5 \cite{dhariya2017hybrid}    & Machine Translation   & 132,300 sentences \\
& & \\
Xlit-Crowd: Hindi-English Transliteration Corpus \cite{khapra2014transliteration}   & Machine Translation   & 14,919 words  \\
& & \\
IIITH Codemixed Sentiment Dataset \cite{prabhu2016towards}  & Sentiment Analysis    & 4,981 sentences \\
\hline
\end{tabular}
\caption{indicnlp\_catalog Hindi-English Datasets}
\label{table:hiEndatasets}
\end{table}

While there is extensive work done in each of these areas, for some reason normalization of Hindi-English (which is at intersection of these areas) hasn't received its due attention. This may be partly due to unavailability of a comprehensive data set and baseline. We believe our work will address some of this.


\section{Corpus Preparation}
While preparing this corpus, we carry out the following steps. 
\begin{enumerate}
\itemsep0em 
    \item \textbf{Data Collection}: collecting Hindi-English sentences.
    \item \textbf{Data Filtering \& Cleaning}: standard pre-processing of raw sentences.
    \item \textbf{Data Annotation}: sentence-level text normalization by human annotators.
\end{enumerate}

\subsection{Data Collection}
We collected data in two phases: In the first phase we built and deployed general chit-chat bots on social media platforms. User responses were randomly sampled and pooled to create the dataset.
In the second phase, we collected data from our platform. Here too the responses were chosen randomly to be added to the dataset.

\subsection{Data Filtering \& Cleaning}
The raw text data we collected was then preprocessed and cleaned. Following were the key steps:
\begin{enumerate}
\itemsep0em 
    \item Drop all messages that were forwarded messages or consisted of only emojis. 
    \item Hindi words were written in both scripts - Devanagari and Roman. All words in Devanagari were converted into roman script.   
    \item Removed all characters other than alpha-numeric characters. 
    \item All sentences containing profane words or phrases were dropped.
    \item All sentences containing any Personal Identification Information (PII) were dropped. 
\end{enumerate}
Steps (4) and (5) were done manually.

\subsection{Data Annotation}
The preprocessed data was sent to human annotators for text normalization annotation. Each word in the input sentence was tagged for the type of non-canonical variation \& its phonetically standard transcription. The annotators chosen were native speakers of Hindi and had bilingual proficiency in English. The dataset was annotated by three annotators while maintaining an average inter-annotator agreement of 95\% on the dataset.

Based on the context of the words in the input sentence, annotators provide the corresponding normalized sentence. Further, to better capture the process used by the annotators to arrive at the normalized text, the annotators provide a unique \textit{tag} for each word. This tag describes the transformation applied by annotators to arrive at the corresponding normalized word. The corpus along with normalized text also contains these tags. Below we describe various tags used in the corpus, the scenario in which a given tag is used and explain the transformation applied with example(s):
\begin{enumerate}
\itemsep0em 
    \item \textbf{Looks Good}: The word under consideration is already an English word with proper spelling e.g. \textit{“yes”}, \textit{“hello”}, \textit{“friend”}
    \item \textbf{Hindi}: The word is a Hindi word in Roman script. In case the spelling is incorrect, replace the word with the corresponding phonetically correct transcription. e.g. \textit{“haaan”} $\rightarrow$ \textit{“haan”}\footnote{Hindi word corresponding to \textit{“yes”} in English }, \textit{“namskar”} $\rightarrow$ \textit{“namaskaar”}\footnote{Hindi greeting corresponding to \textit{“hi”} in English} 
    \item \textbf{Merge}: A word is mistakenly split into two or more consecutive words by uncautious white spaces.  e.g. \textit{“ye s”} $\rightarrow$ \textit{“yes”}, \textit{“hell oo”} $\rightarrow$ \textit{“hello”}, \textit{“fri en dd”} $\rightarrow$ \textit{“friend”}
    \item \textbf{Split}: Two words get conjoined or when a user uses a contraction of two words. Split the words with correct spelling e.g. \textit{“yeshellofriend”} $\rightarrow$  \textit{“yes hello friend”}, \textit{“isn’t”} $\rightarrow$ \textit{“is not”}, \textit{“should’ve”} $\rightarrow$ \textit{“should have”}
    \item \textbf{Short Form}: Word is a short form (phonetically or colloquially). Replace the word with the corresponding full word e.g. \textit{“u”} $\rightarrow$  \textit{“you”}, \textit{“y”} $\rightarrow$ \textit{“why”}, \textit{“doc”} $\rightarrow$ \textit{“doctor”}
    \item \textbf{Acronym}: Word is an Acronym or Abbreviation. Replace with their full form e.g. \textit{“fb”} $\rightarrow$ \textit{“facebook”}, \textit{“brb”} $\rightarrow$ \textit{“be right back”}
    \item \textbf{Typo}: The word is a typo (due to haste, fat-finger error\footnote{https://en.wikipedia.org/wiki/Fat-finger\_error} or low attention to details) while typing e.g \textit{“yass”} $\rightarrow$ \textit{“yes”}, \textit{“helllo00o”} $\rightarrow$ \textit{“hello”}, \textit{“frieendd”} $\rightarrow$ \textit{“friend”}
    \item \textbf{Wordplay}: User has deliberately modified the word for creative purposes. Undo the creativity. e.g. \textit{“hiiiii”} $\rightarrow$ \textit{“hi”}, \textit{“you are punny”} $\rightarrow$ \textit{“you are funny”}, \textit{“pun in ten dead”} $\rightarrow$ \textit{“pun intended”}
    \item \textbf{Profanity}: Word is profane e,g, \textit{“f**k”}, \textit{“ba*tard”}
    \item \textbf{Unrecognizable or other language}: Default tag to catch any non-English/Hindi words or any other type of error/variation not covered by any of the previous tags, like gibberish. e.g.\textit{“\underline{asgjhsad} is my name”}, \textit{“\underline{n233s} is cool”}, \textit{“call \underline{maadi}”}\footnote{Slang that means \textit{“call me”}}
\end{enumerate}

\subsection{Corpus Examples}
To facilitate better understanding, we list out two examples from our dataset to illustrate the input text and corresponding normalized text along with annotations:

\begin{enumerate}
\itemsep0em 
    \item 
    \begin{itemize}
    \itemsep0em
        \item[] \textbf{id:} 5089
        \item[] \textbf{inputText:} \textit{``pakka lisenc nahin hai but ll hai”}
         \item[] \textbf{tags}: ['Hindi', 'Typo’, 'Hindi', 'Hindi', 'Looks Good', 'Acronym', 'Hindi']
        \item[] \textbf{normalizedText:} \textit{“pakka license nahi hai but learning licence hai”}\footnote{Corresponding English translation: ``don't have a permanent license, but I have learning licence”}
       \\
        
        Here, the word 'pukka' is a Hindi word that means `confirmed'. `lisenc' is a typo for `license' and refers to driving license. `nahin' in hindi means `no', so on and so forth. 
        \\
    \end{itemize}
    \item 
    \begin{itemize}
    \itemsep0em
        \item[] \textbf{id}: 13427
        \item[] \textbf{inputText}: \textit{“hiii mjhe jab chaiye”}
        \item[] \textbf{tags}: ['Wordplay', 'Hindi', 'Typo', 'Hindi']
        \item[] \textbf{normalizedText}: \textit{“hi mujhe job chaahie”}\footnote{Corresponding English translation: \textit{“hi, I want a job”}}
        \\
        
        Here, the word 'hiii' is a wordplay for 'hi', 'mjhe' is a typo for hindi word 'mujhe' which means 'I'. 'jab' is a typo for 'job' and `chaiye' is typo for hindi word `chaahie' which means `want'.
        
    \end{itemize}
\end{enumerate}

\section{Corpus Analysis}
\begin{table}[]
\centering
\begin{tabular}{l|c}
\hline
\textbf{}  {\textbf{Attribute}} & \textbf{Value} \\
\hline 
\# Datapoints               & 13494                 \\
\# Train             &   10795        \\
\# Test             &   2699        \\
\% Sentences Modified after Annotation  & 80.08\%   \\
\% Hindi-English Code-Mixing Sentences  & 52.69\%   \\
\% Non-English/Hindi words              & 5.41\%   \\
\% Hindi Words in Corpus                & 41.48\%   \\
Code-Mixing Index (CMI) \cite{das2014identifying}           & 88.40     \\
\hline
\end{tabular}
\caption{Basic Statistics \textit{hinglishNorm} Corpus }
\label{table:basicStats}
\end{table}

After the preprocessing and manual annotation as described in Section 3, we refer to the data set obtained as \textit{hinglishNorm}. It contains 13494 sentence pairs. Table~\ref{table:basicStats} presents some basic statistics of \textit{hinglishNorm} corpus. Each data point in the corpus is a sentence pair consisting of an \textit{inputText} and \textit{normalizedText}. \textit{inputText} is the text as given by the user after preprocessing and \textit{normalizedText} is the corresponding human annotated text. Table~\ref{table:inputTtextVsnormalizedTextStats} gives corpus level statistics of \textit{inputText} and \textit{normalizedText}.

\begin{table}[]
\centering
\begin{tabular}{l|c|c}
\hline
\textbf{Features} & \textbf{\textit{inputText}} & \textbf{\textit{normalizedText}} \\
\hline
\# Sentence  & 13494 & 13494             \\
\# Unique Sentences  & 13066 &  12547    \\
\# Unique Words  & 9326  & 7465          \\
\# Unique Characters & 37 & 37           \\
Most Common Sentence & “whats ur name” & “what is your name”    \\
\# Most Common Sentence & 12 & 38   \\
Mean Character Length & 22.06 & 25.25                   \\
Std Var of Character Length & 16.97 & 19.00             \\
Median Character Length & 18 & 21                       \\
Mean Word Length & 4.96 & 5.13                          \\
Std Var of Word Length & 3.53 & 3.66                    \\
Median Word Length & 4 & 4                              \\
\hline
\end{tabular}
\caption{Statistics for  \textit{inputText} vs \textit{normalizedText}}
\label{table:inputTtextVsnormalizedTextStats}
\end{table}


An important aspect of this corpus is that the \textit{normalizedText} can vary depending on  the context of the sentence. Based on the context of the sentence, misspelled words might require different corrections. For e.g.
\begin{itemize}
\itemsep0em 
    \item \textit{“\underline{hii}, I have a bike” (inputText) $\rightarrow$ “\underline{hi}, I have a bike” (normalizedText)}
    \begin{itemize}
    \itemsep0em 
        \item Input text provided by the user is an English language sentence with misspelled \textit{“hi”}. Annotators understand that the word belongs to English language and correct spelling, in this case, should be \textit{“hi”} 
    \end{itemize}
    \item \textit{“mere pass bike \underline{hii}” (inputText) $\rightarrow$ “mere pass bike \underline{hai}” (normalizedText)}
    \begin{itemize}
    \itemsep0em 
        \item Input text is a romanized version of a Hindi sentence that means \textit{“I have a bike”}. Annotators understand that the word belongs to Hindi language and correct spelling, in this case, should be \textit{“hai”} 
    \end{itemize}
\end{itemize}

\section{Benchmark Baseline}
It is common to model the text normalization problem as a Machine Translation problem\cite{mansfield2019neural}\cite{lusetti2018encoder}\cite{filip2006text}\cite{zhang2019neural}. We built a text normalization model using Bidirectional LSTM with attention on the lines of work by \cite{bahdanau2014neural}. We evaluated our system using well established metrics - Word-Error Rate (WER)\cite{niessen2000evaluation},  BiLingual Evaluation Understudy (BLEU)\cite{papineni2002bleu} and Metric for Evaluation of Translation with Explicit ORdering (METEOR)\cite{banerjee2005meteor}. Table~\ref{table:baselinePerformance} shows the results of our experiments over \textit{hinglishNorm}.

\begin{table}
\begin{center}

\begin{tabular}{|c|c|}
\hline
\textbf{}  \textbf{Evaluation Metric} & \textbf{Baseline} \\
\hline 
WER    &   15.55    \\
BLEU   &   71.21    \\
METEOR &   0.50     \\
\hline 
\end{tabular}
\caption{Baseline Performance on \textit{hinglishNorm}}
\label{table:baselinePerformance}
\end{center}
\end{table}



\section{Conclusion \& Future Work}
We presented \textit{hinglishNorm} version 1.0, a corpus of Hindi-English code mix sentences for text normalization task. Thereby, filling a much needed gap. We provided benchmark baseline results on this corpus. 
In future, we  plan to build stronger baselines like BERT\cite{devlin2018bert} and its variants such as DistilBERT\cite{sanh2019distilbert}, RoBERTa\cite{liu2019roberta}, etc.

\bibliographystyle{unsrt}
\bibliography{hinglishNorm}

\end{document}